\documentclass{bioinfo}
\copyrightyear{2015} \pubyear{2015}

\access{Advance Access Publication Date: Day Month Year}
\appnotes{Manuscript Category}
\usepackage{amsmath}
\usepackage{graphicx}
\usepackage{epstopdf}% To incorporate .eps illustrations using PDFLaTeX, etc.
\usepackage{multirow}
\usepackage{threeparttable}
\usepackage{color}

\begin{document}
\firstpage{1}

\subtitle{Subject Section}

\title[short Title]{Identification of Interaction Clusters Using a Semi-supervised Hierarchical Clustering Method}
\author[Chen \textit{et~al}.]{Yu Chen\,$^{\text{\sfb 1,}*}$, Yuanyuan Yang\,$^{\text{\sfb 1}}$, Yaochu Jin\,$^{\text{\sfb 2}}$ and Xiufen Zou\,$^{\text{\sfb 3,\sfb 4}}$%and Chengwang Xie\,$^{\text{\sfb 3,}*}$
}
\address{$^{\text{\sf 1}}$School of Science, Wuhan University of Technology, Wuhan, 430070, China. \\
$^{\text{\sf 2}}$Department of Computer Science, University of Surrey, Guildford, GU2 7XH,UK.\\
$^{\text{\sf 3}}$School of Mathematics and Statistics, Wuhan University, Wuhan, 430072, China.\\
$^{\text{\sf 4}}$Hubei Key Laboratory of Computational Science, Wuhan University, Wuhan, 430072 China.
%$^{\text{\sf 3}}$School of Computer and Information Engineering, Nanning Normal University, Nanning 530299, China.
}

\corresp{$^\ast$To whom correspondence should be addressed.}

\history{Received on XXXXX; revised on XXXXX; accepted on XXXXX}

\editor{Associate Editor: XXXXXXX}

\abstract{\textbf{Motivation:} Identifying interaction clusters of large gene regulatory networks (GRNs) is critical for  its further investigation, while this task is very challenging, attributed to data noise in experiment data, large scale of GRNs, and inconsistency between gene expression profiles and function modules, etc. It is promising to semi-supervise this process by prior information, but shortage of prior information sometimes make it very challenging. Meanwhile, it is also annoying, and sometimes impossible to discovery gold standard for evaluation of clustering results.\\
\textbf{Results:} With assistance of an online enrichment tool, this research proposes a semi-supervised hierarchical clustering method via deconvolved correlation matrix~(SHC-DC) to discover interaction clusters of large-scale GRNs. Three benchmark networks including a \emph{Ecoli} network and two \emph{Yeast} networks are employed to test semi-supervision scheme of the proposed method. Then, SHC-DC is utilized to cluster genes in sleep study. Results demonstrates it can find interaction modules that are generally enriched in various signal pathways. Besides the significant influence on blood level of interleukins, impact of sleep on important pathways mediated by them is also validated by the discovered interaction modules.\\
\textbf{Availability:} Text  \\
\textbf{Contact:} \href{ychen@whut.edu.cn}{ychen@whut.edu.cn}\\
\textbf{Supplementary information:} Supplementary data are available at \textit{Bioinformatics}
online.
}

\maketitle

\section{Introduction}
Investigation of inherent regulation mechanism is important to discover pathogenic mechanism of abnormity of organism, however,  it is challenging to inferring the disease-related gene regulatory networks~(GRNs), due to the complexity of gene regulatory processes. So, it is widely accepted to study gene regulatory networks (GRNs) by partitioning them into function modules, and then, investigate operation mechanisms of function modules. Thus, it is critical to discover function modules in a correct and efficient way. A lot of approaches have been proposed for module detection based on gene expression data (\citealp{saelens2018comprehensive}). These methods, according to their underlying mechanism, could be divided into two categories:
\begin{enumerate}
\item direct methods identifying interactions between genes, such as GENIE3 (\citealp{irrthum2010inferring}) and TIGRESS (\citealp{haury2012tigress}), etc. These methods generate interaction networks directly, and then, function modules can be obtained according to the generated networks;
\item co-expression methods that generate function modules, including a lot of clustering methods such as FLAME (\citealp{fu2007flame}), WGCNA (\citealp{langfelder2008wgcna}), DBSCAN (\citealp{ester1996density}) and CUBIC (\citealp{li2009qubic}), etc. By identifying similarity among expression profiles, co-expression methods partition together genes with similar expression profiles. In this way, they can get the related function modules.
\end{enumerate}

Direct methods generate an integral network by separating direct connections from indirect ones, which could be misled by data noise and convolution of direct correlation~(\citealp{feizi2013network}). Co-expression methods take some correlation metrics as quantification measure of correlation, and divide genes that are tightly correlated into groups. Compared to the direct methods, they are less sensitive to data noise, and thus, more robust for identification of function modules.
However, although co-expressed genes are usually involved in the same regulatory process, not all genes in a function modules are well co-expressed. For this reason, it is challenging to detect complex function modules, especially when the involved genes are expressed with dissimilar profiles. To improve performance of module detection algorithm, MERLIN (\citealp{roy2013integrated}) tries to combine strengths of both methods by interactively implementing agglomerative hierarchical clustering and probabilistic graphical network; decomposition-based method performs independent component analysis to find independent signals in data, and attribute genes to clusters by the contribution parameters (\citealp{saelens2018comprehensive}).
%
%However, parameters including the total number of independent source signals and the cutoff threshold must be predefined for this method.

A general strategy to improve performances of module detection algorithms is to supervise the discovering process by available information, which leads to various supervised or semi-supervised methods for discovery of function modules. Based on this idea, RNA hybridization images are employed to generate connection constraints to supervise clustering of co-expressed genes (\citealp{costa2007semi}), survival information could be used to preselect interesting features correlated with survival time (\citealp{bair2004semi,koestler2010semi}), and pathogenetic information is incorporated in the double label propagation clustering algorithm to confirm loose connections between genes (\citealp{jiang2017disease}).
For the case that DNA binding information is available, Mishra and Gillies modified the correlation value between nodes connected by DNA binding information, and performed a semi-supervised spectral clustering by the modified affinity matrix (\citealp{mishra2008semi}).

Due to transition of direct connection, correlations between genes are composition of direct and indirect influences, and so,
the convolved correlation measure would lead to disturbed detection results. To separate indirect correlations from direct ones,
we eliminate indirect influence by deconvolution of the correlation matrix (\citealp{feizi2013network}). Meanwhile, if prior information about direct connections are available, we also modify the involved blocks of the correlation matrix, because the available information confirming direct connections between genes should also be transfer to indirect influences, too.  With the modified correlation matrix quantifying direct correlations, the agglomerative hierarchical clustering method is implemented to discover hierarchical structures of GRNs, and then, discover function modules from the hierarchical dendrogram. Furthermore, the proposed method is employed to find influenced function modules by changed sleep settings. We make a first attempt to take gene enrichment results as prior information of supervision, and the results demonstrate that sleep influence not only blood level of various cytokines, but also related signal pathways that have great impact on innate and adaptive immunities of mammalian.

The remainder of this paper is organized as follows. Section \ref{SecMeth} proposes the semi-supervised clustering method based on deconvolution of correlation matrix. In Section \ref{SecRes}, we demonstrate validity of our methods on benchmark networks, and then, employ it to cluster the sleep-related genes in Section \ref{SecSleep} to discover promising results about the mechanism of sleep regulation. At last, the paper is concluded in Section \ref{SecCon}.

\section{Methods}\label{SecMeth}
%In this section, we would like to introduce the details of our proposed semi-supervised hierarchical clustering based on deconvolution (SHC-D).
\subsection{Principles and Challenges of Hierarchical Clustering for GRN Module Detection}
A routine of clustering GRNs by an hierarchical clustering~(HC) method is generally as follows. Given a network $G=(V,E)$ of $n$ vertexes and a data set of $m$ sample points, we first get an $n\times n$ correlation matrix $D(V)$ by some correlation metric. ~\footnote{Without loss of generality, we assume that the diagonal elements of $D(V)$ is $1$, and all non-diagonal elements are located in $(0,1)$. Once a correlation metric is not restricted in $[0,1]$, its value could be standardized to $[0,1]$.}.
Then, the hierarchical clustering is performed to obtain a dendrogram depicting affinity between genes. By cutting the dendrogram via some given threshold $\epsilon$, we can get a clustering results $R_{\varepsilon}(D)$ deriving from a correlation matrix $D$. Besides the general process of HC, one must try to address the following technical issues to get correct results of module detection.
\begin{itemize}
  \item Manage to precisely infer large-scale GRNs with small and noisy data set. Due to expensive expenditure of biological experiments, for most cases available data set is of small size. Meanwhile, these experimental data are usually noisy;
  \item Reveal complete interactions of genes by gene expression profiles. Although genes with similar expression profiles often belong to shared function modules, a function module could include genes that are not tightly co-expressed. So, some members could be missed if modules are clustered via expression-oriented correlation metrics;
  \item Mining prior information to supervise HC when extra experimental information is unavailable for improvement of module detection.
\end{itemize}
Our study, in this paper, tries to dispose these problems as follows.
\begin{itemize}
\item With respect that available data sets are usually small and noisy, we take the agglomerative hierarchical clustering (AHC) method as the tool of module detection. Because AHC is compatible to the data noise, its robustness can be regulated by appropriate setting of the cutting threshold;
\item In order to supervising the clustering results, we would like to modify the correlation matrix by available prior information. In this way, a semi-supervised AHC is proposed. For the first time we take the gene enrichment results as prior information, and it is expected to combine together genes that are loosely co-expressed but contribute to shared function. Meanwhile, it could also distinguish genes that are expressed with similar profiles but belong to different modules;
\item Because incorporation of prior information could increase average correlation values of a neighborhood, we take the single-linkage method to generate clusters, in case that the increased indirect correlation mislead the cluster results. Meanwhile, we also separate indirect dependence by deconvolving correlation matrices~\citep{feizi2013network};
\end{itemize}

\subsection{Incorporation of Prior Information}\label{IPI}

When we try to discover interaction modules of GRN by the expression profiles, available prior information  could be depiction of gene regulations, part of protein-protein interaction network, and part of information transduction pathways, etc., which are either a collection of known direct interactions or a set of known communities. Suppose that available information says that some genes in the subset $V_S=\{v_{s_1},\dots,v_{s_k}\}$ have some mutual interactions, denoted as $E_S=\{e_{s_1},\dots,e_{s_l}\}$. If the subnetwork $G_{S}=(V_S,E_S)$ is connected, we get a known community $C=V_S$; if the disconnected subnetwork $G_{S}=(V_S,E_S)$ consists of $p$ connected $G_{1}=(V_{S_1},E_{S_1}),\dots,G_{p}=(V_{S_p},E_{S_p})$, we get $p$ known communities $V_{S_1},\dots,V_{S_p}$. So, no matter which kind of prior information is available, we can translate it into a collection of known communities. In order to discriminate them from the final clusters results, we name these communities derived from available prior information as \emph{prior clusters} in the rest of this paper.

Denote $C_1,C_2,\dots,C_m$ as prior clusters certificated by prior information, where size of $C_k$ is $n_k,\,k=1,\dots,m$. We can obtain a $n_k\times n_k$ sub-matrix of correlation for each cluster $C_k$, denoted as $D(C_k)$.
Promoting to cluster together genes in $C_k$, we would like to strengthen dependencies between genes by increasing all elements of $D(C_k)$ to the same extent. Motivated by this idea, non-diagonal elements of the sub-matrix $D(C_k)$ is divided by a parameter $\gamma_k$. Denote
\begin{align*}
 & D(C_k)  =(d_{i,j}(C_k))_{n_k\times n_k}, \\
 & d(C_k)  =\max_{i\neq j}d_{i,j}(C_k).
\end{align*}
Non-diagonal elements of $D(C_k)$ can be magnified as
\begin{equation}\label{D_M}
 d'_{i,j}(C_k)=\left\{\begin{aligned} & d_{i,j}(C_k)/\gamma_k, && \mbox{ if } i\neq j,\\ & d_{i,j}(C_k) && \mbox{ if } i\neq j,\end{aligned}\quad i,j=1,\dots,n_k,
 \right.
\end{equation}
where $\gamma_k=\rho d(C_k)$, $\rho\in[1,1/d(C_k)]$.
Since $d(C_k)$ is restricted in $(0,1)$, $\gamma_k$ is valued in $(0,1)$, too. Replacing submatrix $D(C_k)$ by $D'(C_k)=(d'_{i,j}(C_k))_{n_k\times n_k}$, we can tighten this enriched cluster and get an updated correlation matrix $D'(V)$. Performing AHC via $D'(V)$, we carry out a semi-supervised clustering for the investigated gene list $V$.

The parameter $\rho$ reflects reliability of the prior information. $\rho=1$ implies that the prior information is correct for sure, and so, the maximum correlation $d(C_K)$ should be magnified to $1$; on the contrary, if the prior information is totally useless, the original correlation matrix should keep unchanged, which is realized by setting $\rho=1/d(C_k)$ to make $\gamma_k=1$.

If prior information provides multiple enriched clusters for the investigated network, there are two different ways to incorporate the prior information, named as the ``global incorporation'' and the ``local incorporation'', respectively.  The ``global incorporation'' implement  the magnification process introduced by (\ref{D_M}) for all enriched clusters, and one globally updated correlation matrix is obtained; however, the ``local incorporation'' only incorporates one enriched cluster at one time, and consequently, multiple ``locally'' updated correlation matrices are obtained.

\subsection{Semi-supervised Hierarchical Clustering via the Deconvolved Correlation Matrix}\label{HCDCM}
Due to transmission of regulation, mutual correlation between two gens could be composition of direct and indirect regulations. Thus, identification precision of interaction clusters greatly depends on separation of direct regulation from the composed oneps. \cite{feizi2013network} proposed a  mathematical method named as the ``network deconvolution'' to separate the direct correlation from observed correlations. A correlation matrix $D_{obs}$, called as an observed correlation matrix, is regarded as composition of the direct correlation matrix. Then, the observed correlation matrix could be decomposed as follows.
\begin{equation}\label{CorDec}
D_{obs}=D_{dir}+D_{indir}=D_{dir}+\sum_{n=2}^{+\infty}D_{dir}^n=D_{dir}\sum_{n=0}^{+\infty}D_{dir}^n,
\end{equation}
where $D_{dir}$ is the direct correlation matrix, and $D_{indir}$, the indirect correlation matrix, represents all-order combination of all indirect correlations. Once the observed correlation matrix $D_{obs}$ is linearly scaled so that all eigenvalues of $D_{dir}$ are between $-1$ and $+1$, the right-hand side of equation (\ref{CorDec}) is equal to $D_{dir}\left(I-D_{dir}\right)^{-1}$, and we have
\begin{equation}\label{Gdir}
    D_{dir}=D_{obs}\left(I+D_{obs}\right)^{-1},
\end{equation}
whichp could be efficiently computed  via the singular value decomposition (SVD) method (\citealp{feizi2013network}).

Then, a semi-supervised hierarchical clustering method via deconvolved correlation matrix~(SHC-DC) is proposed for detection of function modules. Given a correlation metric and a gene lists $V$ with corresponding expression profiles data, it first computes correlation matrix $D(V)$ for the gene list $V$. Then, modified correlation matrix $D'(V)$ is obtained by incorporating available prior information to $D(V)$. Extracting the direct correlation matrix $D'_d(V)$ by equation (\ref{Gdir}) from $D'(V)$ , SHC-DC performs the agglomerative hierarchical clustering \citep{rocach2005clustering} to construct a dendrogram identifying mutual dissimilarities between genes, where  the 'single' linkage method is used to construct the tree-like connections, because the clustering process is based on a correlation matrix of direct correlations.

\section{Performance Demonstration of SHC-DC on Benchmark Networks}\label{SecRes}

\subsection{Correlation Metrics}
Identification of function modules is based upon quantification of mutual interactions between genes, which is realized by various correlation metrics. Due to the inherent mechanisms of gene regulation, we know that correlations between genes, if direct regulation exists, should be nonlinear. Meanwhile, sometimes their correlation is linear, such as the co-expressed genes that are regulated in similar mechanisms.  Thus, we construct the observed correlation matrices by the Pearson Correlation Coefficient~(PCC), the Distance Correlation Coefficient~(DCor)~(\citealp{szekely2007measuring}) and the Mutual Information~(MI). For computation of MI between real-valued vectors, the empirical estimator of Paninski (denoted as MI1) (\citealp{paninski2003estimation}) , the empirical estimator corrected by Miller-Madow (denoted as MI2) (\citealp{paninski2003estimation}) and the shrink entropy estimator (denoted as MI3) (\citealp{schafer2005shrinkage}) are utilized to obtain approximations of MI, all of which are implemented by the R/bioconductor \emph{minet} package (\citealp{meyer2008minet}). %Then, benchmark problems are investigated via the observed correlation matrices, and then, we employ it  clustering genes related to sleep.

\subsection{Comparison of Performances for Benchmark Problems}

Before application of SHC-DC on module detection of real networks, we first validate its performances by comparing its performances on benchmark networksp with existing clustering algorithms.

\subsubsection{Networks, Expression Data and Module Definitions}

Performance comparisons are based on the benchmark problems proposed for the DREAM5 network inference challenge~(\citealp{marbach2012wisdom}), including an \emph{E. coli} network and two \emph{Yeast} networks. Gene expression data of the \emph{E. coli} and \emph{Yeast} networks are downloaded from the DREAM5 network inference challenge website\footnote{{https://www.synapse.org//\#!Synapse:syn2787209/wiki/70349}}, %(synapse.org/\#!Synapse:syn2787209/wiki/70349),
and the ``observed'' correlation matrices are generated via PCC, DCC, MI1, MI2 and MI3, all of which are implemented by the R/bioconductor \emph{minet}  package~(\citealp{meyer2008minet}).%, denoted as $G_P, G_D, G_{M1}, G_{M2}, G_{M1}$, respectively.

Evaluation of clustering results depends on some gold standard results of gene clusters.  The \emph{E. coli} network is downloaded from the RegulonDB database version 8\footnote{{http://regulondb.ccg.unam.mx (accessed on 03/06/2015)}}, and two \emph{Yeast} networks are respectively generated from an integration of chromatin immunopurification-on-chip data and conserved binding motifs (\citealp{macisaac2006improved}) as well as a combination of genome-wide transcription factor binding data, knockout expression data and sequence conservation (\citealp{ma2014novo}). As reported by \cite{marbach2012wisdom}, gold standard clusters are generated by known network structures as follows~\citep{saelens2018comprehensive}.
\begin{enumerate}
  \item \emph{Minimally co-regulated modules}, denoted as \emph{`minimal'} for short, are defined as overlapping groups of genes that shared at lest one regulator;
  \item \emph{Strictly co-regulated modules}, denoted as \emph{`strict'} for short, are defined as groups of genes known to be regulated by exactly the same set of regulators;
  \item \emph{Strongly interconnected known modules} are defined as group of genes that are strongly interconnected. Three graph clustering algorithms are employed to generate strongly interconnected known modules. For the  Markov clustering algorithm~(\citealp{van2000graph}), inflation parameters 2, 10 and 50 are taken to generate three gold standards, named as \emph{`mc1'}, \emph{`mc2'}, \emph{`mc3'}, respectively. The transitivity clustering~(\citealp{wittkop2010partitioning}) generates two gold standards \emph{`tc1'} and \emph{`tc2'} by setting cutoff parameters for the fuzzy membership 0.1 and 0.9. For the affinity propagation~(\citealp{frey2007clustering}) we varied the preference value between 0.5,2 and an automatically estimated value to generate three gold standards \emph{`ap1'}, \emph{`ap2'} and \emph{`ap3'}.
\end{enumerate}

%{\color{red} By the Markov clustering algorithm (\citealp{van2000graph}), the transitivity clustering (\citealp{wittkop2010partitioning}) and the affinity propagation (\citealp{frey2007clustering}), we generate for each network three gold standard results, which respectively represent the minimally co-regulated modules, the strictly co-regulated modules as well as the strongly interconnected known modules. Something Wrong!!!!}
The co-regulated modules are likely to consist of co-expressed genes, while the strongly interconnected known modules are clustered by their direct regulatory relation. Various gold standards could be regarded as different ``interpretations'' of the corresponding GRNs. Inspired by this idea, we also take the known gold standard clusters as some kind of prior information. When the gold standard result coincides with the prior information, we get precise prior information about interactions between genes; otherwise, it means that the known prior information include some incorrect contents.

%Moreover, incorporation of the ``prior information'' is also performed for the direct correlation matrices generated by the deconvolution method demonstrated in Section \ref{HCDCM}. Then, for each network we get two correlation matrices generated by a correlation metric: one consists of the original metric values, and the other is the corresponding deconvolved matrix that only depicts direct correlation. Then, comparison among results obtained based on each correlation matrix can demonstrate the influence of correlation deconvolution and composed effect of deconvolution and supervision.

\subsubsection{Numerical results and discussion}
Then, for three investigated networks (denoted as \emph{E. coli}, \emph{Yeast1} and \emph{Yeast2}), we can get ten dependency matrices, including five correlation matrices by PCC, DCC, MI1, MI2 and MI3 and  five corresponding matrices quantifying the direct correlation, which are generated by the deconvolution method \citep{feizi2013network}. For totally ten correlation matrices, we can detect function modules by the AHC and the SHC-DC supervised respectively by 11 set of prior information. In a word, we can get totally 120 group of clustering results for each network.

Then, 120 clustering results are respectively compared with 11 gold standard results  for evaluation. Since each clustering result is a dendrogram, by sampling cutting thresholds from the minimal distance to the maximal distance for all obtained dendrograms, we can collect the number of True Positive~(TP), False Positive~(FP), True Negative~(TN) and False Negative~(FN), and generate 11 receiver operation curves (ROCs) for each clustering. Then, the area under curve (AUC) is employed to assess integral performance of each result, and the F Measure is used to evaluate the best result for each ROC. For each network, heatmaps of AUC are included in Figs. 1-3 in the Supplementary Material. For each combination of clustering method, correlation metric and network, Tab. \ref{Tab1} includes the AUC and F measures averaged across all combinations of prior information and gold standard.

\begin{table}[!htp]
\caption{Comparison of Overall Performances on Benchmark Problems for AHC, AHC-D, SHC and SHC-DC. \label{Tab1}}
\centering
\begin{tabular}{ccccccccc}
  \hline\hline
   \multirow{2}{*}{Method} & \multirow{2}{*}{Metric} &\multicolumn{3}{ c }{ AUC}  & & \multicolumn{3}{ c }{ F-Measure}   \\
   \cline{3-5} \cline{7-9}
    & &\emph{E. coli}   & \emph{Yeast1}  & \emph{Yeast2} & & \emph{E. coli}  & \emph{Yeast1} & \emph{Yeast2}  \\
    \hline
    \multirow{5}{*}{AHC}&DCC&0.55&0.51&0.50& &0.052&0.019&0.005\\
    &PCC&0.55&0.51&0.48& &0.057	 &0.019&0.008\\
    &MI1&0.57&0.52&0.55& &0.041 	 &0.014&0.003\\
    &MI2&0.57&0.52&0.54& &0.043 	 &0.016&0.004	\\
    &MI3&0.57&0.52&0.55& &0.043  &0.015&0.003\\
    \hline
    \multirow{5}{*}{AHC-D}&DCC&0.64&0.55&0.63& &0.112&0.042 &0.074\\
    &PCC&0.65&0.55&	0.63& &0.123&0.032&0.046\\
    &MI1&0.60&0.52&	0.57& &0.067&0.019&0.010\\
    &MI2&0.59&0.53&	0.57& &0.071& 0.022&0.015\\
    &MI3&0.59&0.52&	0.57& &0.072&0.019&0.010\\
    \hline
    \multirow{5}{*}{SHC}&DCC&0.57&0.56&0.58& &0.077&0.041&0.053\\
    &PCC&0.57&0.55&0.56& &0.080&0.041&0.050\\
    &MI1&0.53&0.53&0.58& &0.025&0.014&0.003\\
    &MI2&0.54&0.55&0.59& &0.029&0.018&0.004\\
    &MI3&0.54&0.53&0.57& &0.027&0.015&0.003\\
    \hline
    \multirow{5}{*}{SHC-DC}&DCC&0.66&0.63&0.70& &0.130&0.075&0.140\\
    &PCC&0.66&0.64&0.71& &0.138&	0.063&0.108\\
    &MI1&0.57&0.55&0.60& &0.044&	0.023&0.016\\
    &MI2&0.57&0.57&0.63& &0.054&	0.030&0.023\\
    &MI3&0.57&0.55&0.61& &0.048&	0.024&0.016\\
  \hline
\end{tabular}
\end{table}
Comparison of AUC values demonstrates that the nonlinear metric MIs (including three approximation estimator, MI1, MI2 and MI3) are generally more appropriate than the linear metric PCC and DCC for the AHC algorithm. MIs better quantify the composed nonlinear relation between genes than PCC and DCC, and so, for most of the networks MIs lead to better overall performances of the single linkage method. However, the results on (best) F-measure demonstrate a contrary result. Although nonlinear metrics (MI) can get better overall performance, it is also possible to get better results via the linear metrics (DCC \& PCC), once we can choose appropriate parameters to optimize performance of clustering algorithms. This could be the potential reason why linear correlation metrics are widely employed regardless of the underlying nonlinear correlation between genes.

%Cases that linear metrics perform better  could be attributed to the properties of gold standard clusters. If they are generated with preference to clustering co-expression genes together, the evaluation results corresponding linear correlation metrics could be raised greatly.

Compared with AHC, AHC-D performs much better on the benchmark problems, which is attributed to introduction of deconvolution~\citep{feizi2013network}. Contrary to the case of AHC, for AHC-DC linear correlation metrics (DCC \& PCC) work better than the nonlinear MIs.

The observed correlation includes transitive indirect correlation, which is transmitted in a system way, and so, could not correctly display direct correlation between genes. In comparison, the deconvolved DCor and PCor matrices are good approximations of direct mutual dependencies, and so, result in better performance when we use hierarchical clustering methods with the single linkage. However, the promising function of deconvolution is not significant for three numerical MI indexes, because the deconvolution routine is performed with the underlying assumption that correlation is linearly composed, which does not hold for the nonlinear metric MI. When the observed MI matrices are deconvolved according to (\ref{Gdir}), the obtained results are not nice quantifications of direct correlations. Meanwhile, worse results for MI metrics could be caused by numerical approximation of MI metrics, because expression profiles must be aligned to discrete bins when MI between genes is computed by the aforementioned estimators. Then, loss of information could be another cause of worse performance of MI for the AHC-D.

By incorporating prior information, the clustering results are further improved. Increases of AUC and F-measure imply that both overall performance and optimal performance of SHC-DC are significantly better than AHC-D. Recall that various gold standards for the same network are generated from the same network topology according to different principles, and so, they have some modules structure in common, and simultaneously, some differences that can be taken as a kind of inaccuracy of prior information. Although the ``inaccuracy'' exists, results in Tab. \ref{Tab1} demonstrates their incorporation is generally effective. Meanwhile, the heatmaps of Fig. 1-3 in the supplementary materials also show that whether or not prior information coincides with the gold standard, the semi-supervision by introduction of  prior information is very promising for improvement of clustering results.

The final clustering result is obtained by choosing one point from the ROC curve, which is purpose-dependent dependent on the evaluation metric of clustering results. The chosen point could be a point maximizing some external metric, or a point located on a special position of the ROC curve. In this paper, we take the point with the maximum F-measure as the clustering result, and evaluate it via the Precision, Recall and F-measure (\citealp{saxena2017review}). Noting that both DCC and PCC outperform MIs in our studied cases, we only compare SHC-DC for DCC and PCC with the popular clustering method WGCNA (\citealp{langfelder2008wgcna}) and QUBIC (\citealp{li2009qubic}). The overall performance of our method, for each correlation correlation (DCC or PCC), is evaluated by averaging metric values across all settings of prior information and gold standards, and compared with the results of WGCNA and QUBIC in Tab. \ref{FTable}.

\begin{table}[!htp]
\caption{Performance Comparison between SHC-DC and Popular Clustering Methods. \label{FTable}}
\begin{tabular}{ccccc}
\hline\hline
{Method}&{Network}&Precision&Recall&F-Measure\\

\hline
\multirow{3}{*}{WGCNA}&Ecoli&0.0133 &	0.3241 &	0.0247\\
& Yeast1 & 0.0161 &	0.1544 &	0.0277\\
& Yeast2 & 0.0029 &	0.2194 &	0.0057\\

\hline
\multirow{3}{*}{QUBIC}&Ecoli&0.0266 &	0.0807 &	0.0331\\
& Yeast1 & 0.0188 &	0.0416 &	0.0228\\
& Yeast2 & 0.0022 &	0.0441 &	0.0041 \\

\hline
\multirow{3}{*}{SHC-DC/DCC}&Ecoli&0.1621 &	0.1269 & 0.130	 \\
& Yeast1 & 0.1044  & 0.0651	  &	0.075 \\
& Yeast2 & 0.1966  & 0.1173	  &	 0.140\\

\hline
\multirow{3}{*}{SHC-DC/PCC}&Ecoli&0.1693 &	0.1364 & 0.138	\\
& Yeast1 & 0.0850 &	0.0558 & 0.063	\\
& Yeast2 & 0.1377 &	0.0986 & 0.108	\\
\hline\hline
\end{tabular}
%\begin{tablenotes}
%\item[xxxx] Please note that the results of MyMethod is averaged for all prior information. The Precision, Recall and F Measure are averaged respectively.
%\end{tablenotes}
\end{table}

Results indicates that SHC-DC is generally superior to the WGCNA and QUBIC, for both DCC and PCC metrics, which demonstrates the superiority of our semi-supervised clustering method, even when the ``prior information'' is always inaccurate. For details, the PCC metric  performs a bit better than the DCC metric on the \emph{Ecoli} network, and worse on the \emph{Yeast} networks. Recall that the DREAM5 data are generated with artificial models based on known network topologies. Then, the results could be attributed to better accommodation of PCC metric on correlations between genes, if available data are sufficiently precise for accurate identification of network modularity. On the contrary, available data cannot provides sufficient information of the greater \emph{Yeast} network, and it is anticipated that the DCC metric performs better on it than the PCC .

\section{Clustering Genes Related to Sleep}\label{SecSleep}
To demonstrate the potential application value of SHC-DC in bioinformatics, we employ it clustering sleep-related genes.  It has been reported that chronic sleep loss, total deprivation of sleep, and mistimed sleep influence genomic expression of mammalian (\citealp{moller2013effects,archer2014mistimed,arnardottir2014blood}). When sleep settings changed, expression profiles of genes could keep unchanged, or change from circadian states to in-circadian states, and vice versa. However, the underlying influence mechanism is not clear. Regulation mechanism of sleep conditions could be discovered by clustering sleep-related genes across all sleep settings. If it is possible to identify coarse interaction modules of these, precise regulation mechanism could be further inferred. However, due to the wide influence of sleep conditions, these genes could be expressed with different profiles, and it is necessary to supervise the clustering process by available prior information.

According to the  published researches \citep{moller2013effects,archer2014mistimed}, our gene list includes genes that are reported circadian for sleep in phase (IP), sleep out of phase (OP), total sleep deprivation (TSD) after sufficient sleep (SS) and TSD after insufficient sleep (IS), as well as genes that are reported to be mainly effected by sleep conditions.  This list contains 4893 items, including some probes that  are presently not attributed to known genes~(See the Supplementary table for the detailed gene list).

\subsection{Merging Genes with Similar Expression Profile by a Strict Threshold}
We first merge co-expressed genes to meta genes. This operation can not only reduce total number of investigate genes, but also combine co-expressed genes together to reduce the side effect of semi-supervising---once only some of these co-expressed genes are included in the ``prior information'', they could be separated by markedly magnified difference on the modified correlation values incorporating supervision information. Note that mergence of these co-expressed genes is not confusing when they belongs to difference interaction modules, because genes with very similar expression profiles are almost necessarily regulated in the same way, and it is reasonable to merge them into one metagenes, which then belongs to all related modules. However, this combination should be performed in a very strict way. Thus, we take the PCC as similarity of gene expression profiles, and then generate meta-genes by merging genes with PCC values greater than $0.95$. Finally, totally 4803 meta-genes are generated~(See the Supplementary table for the detailed gene list), where expression profiles of meta-genes are averaged across all member genes.

\subsection{Clustering Meta-Genes using Semi-supervised Hierarchical Methods}

\begin{table}[!htp]
\caption{AUC Values of the ROC Curves Generated by Clustering the Sleep-related Genes.}\label{AUCTable}
\begin{tabular}{ccccccc}
\hline\hline
\multirow{2}{*}{Correlation}&\multirow{2}{*}{Prior Inf.} & \multicolumn{2}{c}{Undeconvoved}& & \multicolumn{2}{c}{Deconvolved}\\
\cline{3-4} \cline{6-7}
& & BP &PW& &BP &PW\\
\hline
\multirow{3}{*}{DCC} &Null & 0.555 & 0.583& & 0.567 & 0.594\\
&BP & 0.733 & 0.647& & 0.779 & 0.683\\
&PW & 0.593 & 0.694& & 0.628 & 0.777\\

\hline
\multirow{3}{*}{PCC} &Null & 0.561 & 0.594& & 0.562 & 0.579\\
&BP & 0.696 & 0.589& & 0.745 & 0.626\\
&PW & 0.641 & 0.675& & 0.660 & 0.752\\

\hline
\multirow{3}{*}{MI1} &Null & 0.559 & 0.588& & 0.563 & 0.597\\
&BP & 0.560 & 0.587& & 0.574 & 0.603\\
&PW & 0.578 & 0.625& & 0.587 & 0.643\\

\hline
\multirow{3}{*}{MI2} &Null & 0.556 & 0.589& & 0.562 & 0.597\\
&BP & 0.611 & 0.610 & &0.629 & 0.625\\
&PW & 0.583 & 0.647 & &0.594 & 0.664\\

\hline
\multirow{3}{*}{MI3} &Null & 0.554 & 0.588& & 0.561 & 0.596\\
&BP & 0.574 & 0.594& & 0.591 & 0.609\\
&PW & 0.578 & 0.636& & 0.590 & 0.652\\

\hline\hline
\end{tabular}
%\begin{tablenotes}
%\item[xxxx] xxxxxx.
%\end{tablenotes}
\end{table}

Mutual correlation between Meta-genes is  evaluated via the PCC, the DCC as well as three numerical estimators of the mutual information denoted as MI1, MI2 and MI3. For each correlation metric, we perform hierarchical clustering based on the original correlation matrices, the deconvolved correlation matrices, and the deconvolved correlation matrices updated by prior information. Because online enrichment analysis tools can provide collected information on given gene lists, in this study we incorporate the online enrichment analysis results as prior information for supervision.  By uploading the full list of all 4893 items to Metascape~\citep{zhou2019metascape}, we perform the Biological Process (BP) enrichment and the Canonical Pathways (PW) enrichment, and take the respective enrichment results as two set of prior information. Then, two clustering dendrograms are obtained by hierarchical clustering. Since there is no real gold standard for the investigated network, we also take the enrichment results as the gold standard results for evaluation of clustering results. Note that the prior information and the gold standard clusters are not always coincident. By sampling single linkage values of correlation from the minimal mutual correlation to the maximal mutual correlation with sampling step $0.01$, we can obtain ROCs and the corresponding AUC values that are included in Tab. \ref{AUCTable}, where ``Prior Inf.'' represents the prior information used for supervision, ``Undeconvolved'' means that the correlation matrices are not deconvolved, and ``Undeconvolved'' indicates that the correlation matrices as deconvolved via formula (\ref{Gdir}). When the ``Prior Inf.'' is ``Null'', it means the clustering process is not supervised by prior information.

Generally, both supervision and deconvolution improve the clustering results. The improvements are especially significant again when we employ the linear metrics PCC and DCC, even if the gold standard and prior information are not coincident. Because the convolution process is linear, linear metrics PCC and DCC can be deconvolved precisely. As a result, deconvolution of PCC and DCC values could be a nice approximation of direct correlation between genes.  The reason that MI cannot get satisfactory results is attributed to the irregular data noise and small size of data set. For this case, simulation results of MI cannot indicate the precise nonlinear correlation between genes. On the contrary, linear PCC and DCC can approximate it with satisfactory precision, and so, perform better.

It is noted that DCC performs a bit better than PCC. Because the distance correlation only evaluate linear correlation of order sequence between two arrays, it only depends on the monotonicity of values. Maybe, this mechanism could greatly improve its robustness to data noise, and to some extent, incorporate the nonlinear relation between gene expression profiles.

%For comparison, we found that DCor and PCor generally perform better than Mutual Information, for all 3 different estimation methods. Most empirical studies show that for most cases MI could not perform better than PCor, which is also consistent with this study. It is attributed to the fact that the direct mutual information between genes, as a kind of nonlinear metric, could not be correctly separated by deconvolution. When the observed MI matrices are deconvolved, the obtained results could be a worse quantification of direct correlation. Moreover, worse results for MI metric is caused by numerical approximation of MI metrics. When MI between genes is computed by the aforementioned estimator, expression profiles must be aligned to discrete bins, which leads to significant loss of information. However, Pcor, although only quantifies linear relation, could be an approximation of nonlinear correlation. According to the Tailor's theorem, this approximation could quantify mutual correlation if the nonlinear part is minor, which is actually the case because mutual function could be modeled by the Hill's function. Moreover, Dcor is a linear metric that can to some extent identify the nonlinear correlation, and so, performs better than PCor.

\subsection{Biological Analysis of Clustering Results}
Incorporation of prior information could connect genes that are biologically related, which can lead us to a global view to the regulation process. Consequently, we can cluster together genes that belong to the same function modules but have different expression profiles. Because the ``biological process'' and ``pathway'' enrichment results contain respective biological significance, we analyze the clustering results based on the DCC matric supervised by the ``biological process'' enrichment and the ``pathway'' enrichment, respectively. Note that the following results are obtained by taking gold standards different from the prior information. That is, when we take the BP enrichment results as the prior information, the result is evaluated by the PW enrichment result; while the PW enrichment result is take as the prior information, we evaluate the clustering result by the BP enrichment result.
\begin{figure}[!htp]
\centering
\includegraphics[width=3.2in]{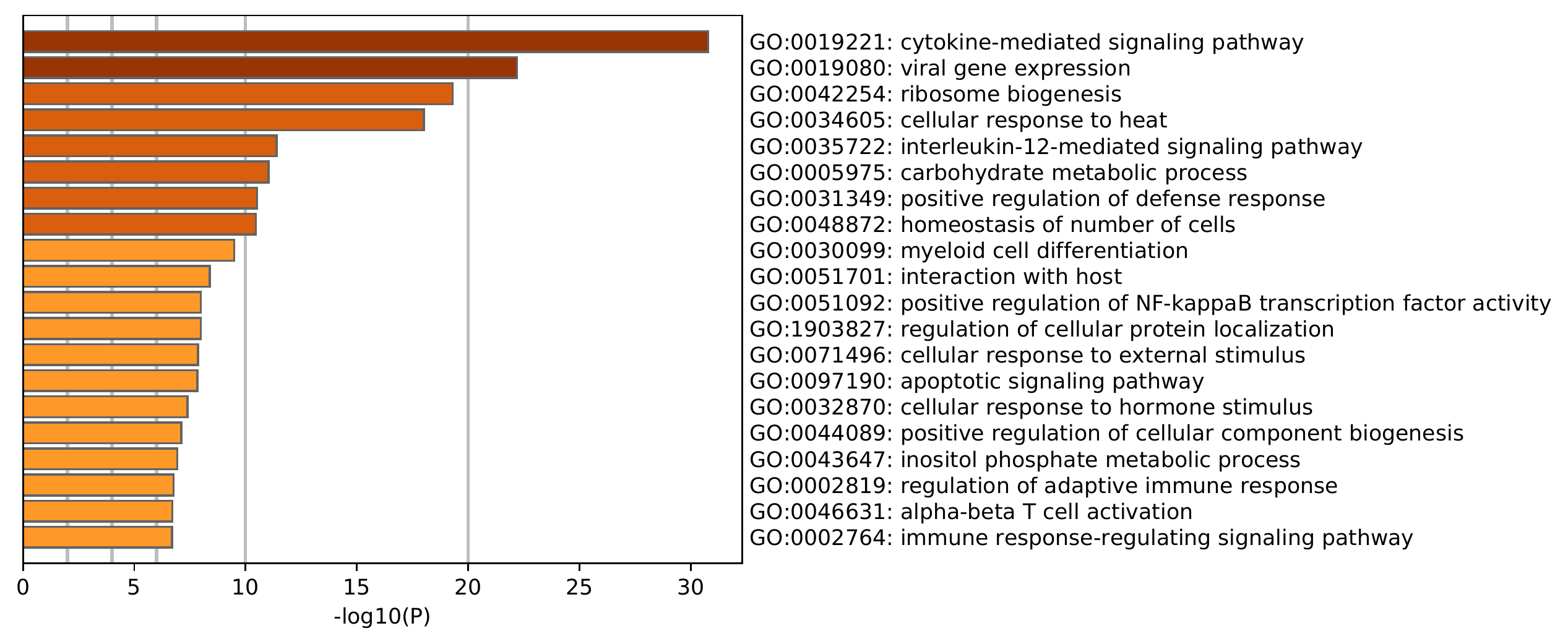}
\caption{Bar graph of BP-enriched terms across input gene lists. Genes are clustered with supervision of PW enriched prior information, and evaluated by BP-enriched gold standard.}
\label{fig_1}
\end{figure}

\begin{figure}[!htp]
\centering
\includegraphics[width=3.2in]{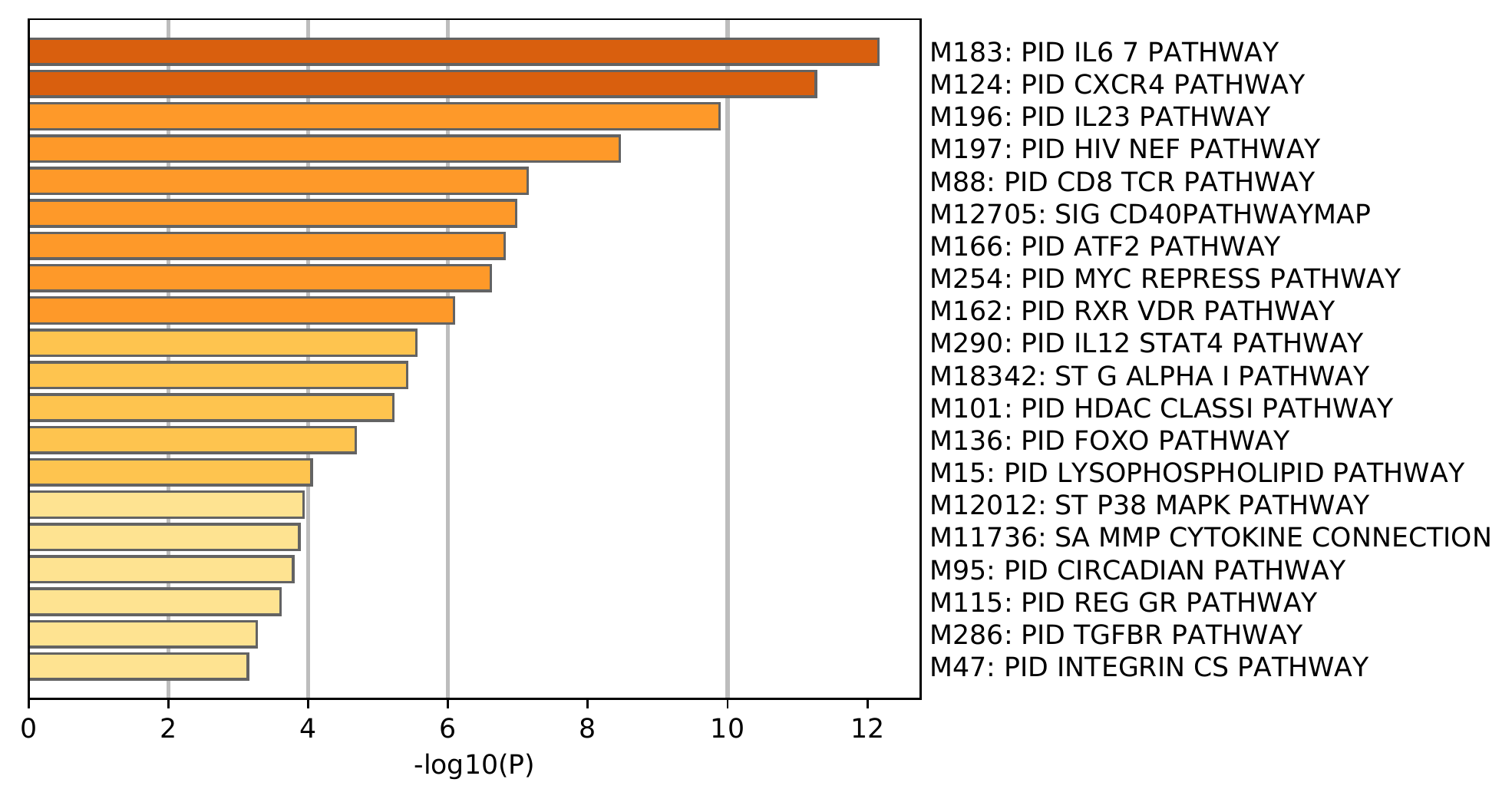}
\caption{Bar graph of PW-enriched terms across input gene lists. Genes are clustered with supervision of BP-enriched prior information, and evaluated by PW-enriched gold standard.}
\label{fig_2}
\end{figure}

As GRNs are still very incomplete~\citep{R2012How}, there could be a strong overestimation of the number of false positives in the observed modules, as most of these genes probably belong to one or more co-regulated modules, which we do not yet know~\citep{saelens2018comprehensive}. Thus, it could be biased if we choose the final result via metrics depending on computation of FN, and we locate the knee area on the ROC curves and take the corresponding thresholds to get the final clustering results from the clustering dendrograms. Because the ROC curve is located in $\mathbb{R}^2$, we use the angle-based method~\citep{branke2004finding} for seeking knee points on the ROC curves.

A final clustering result consist of multiple obtained clusters. To focus on the most important cluster, we take genes in the cluster of the maximum size as the investigated genes. Because our clustering results are based on combined expression profiles under varied sleep settings, the maximum cluster could be a collection of the most influenced genes. Uploading the gene list to the Metascape online tool, we perform enrichment analysis by the categories identical to those of the gold standards.

When the SHC-DC algorithm takes the PW enrichment result as prior information and evaluates the clustering results via the BP enrichment result, the top 20 enriched modules sorted by the logarithm of p-values  are included in Fig. \ref{fig_1}. It has been reported that sleep does influence the blood levels of various cytokines like IL-1, IL-2, IL-6, IL-7, IL-12, TNF-$\alpha$ and IFN-$\gamma$, and assumed that the innate and the adaptive immune functions of mammalian are in turn influenced~\citep{Luciana2012Sleep}, which is here validated by the enrichment results of the clustering results. It is indicated that the obtained gene list, which could be the collection of genes significantly influenced by sleep settings, is enriched for the biological processes such as the ``cytokine-mediated signalling pathway'', and the ``interleukine-12-mediated signalling'', etc. Then, we can conclude that influence of sleep on immunity is achieved not only by regulating blood level of various cytokines, but also by adjusting the related signal transduction pathways. What is interesting is that the ``ribosome biogenesis process'' is also enriched for the clustered results, which maybe reveal the inherent mechanism of wide influence of sleep on genome-wide expression: general impact of sleep on the ribosome biogenesis influences general translation of RNA to proteins, and then, affect expression of a large number of genes by regulating generation of most transcription factors.

Meanwhile, we also run the SHC-DC algorithm with supervision of the BP-enrichment result and evaluate the clustering results via the PW-enrichment result.  The top 20 enriched modules sorted by the logarithm of p-values  are included in Fig. \ref{fig_2}. The first enriched pathway is the ``PID IL6 7 pathway'', which shows again sleep influences not only the blood level of interleukins IL-6 and IL-7, but also the related signal pathways, by which the immunity system is then controlled.

\section{Conclusion}\label{SecCon}
%\vspace*{-10pt}
In order to detect function modules consisting of genes with various expression profiles, this paper proposes a semi-supervised hierarchical clustering method via deconvolved correlation matrix~(SHC-DC). By comparing the performance of this method with the popular methods WGCNA and CUBIC, clustering results on the benchmark \emph{E. coli} and \emph{Yeast} networks demonstrate that the semi-supervision of prior information does help to locate genes belonging to the same function modules, and the linear correlation metrics with assistance of correlation deconvolution can generally perform better than the nonlinear metrics. Application of SHC-DC on the sleep-related gene list also shows applicability of SHC-DC: the results indicate that regulation of sleep on immunity of mammalian is achieved by influencing not only the blood level of cytokines, but also the signalling pathways that are mediated by them to activate the immune process.

%\section*{Acknowledgements}

%\section*{Funding}
%
%This work was supported by the National Natural Science Foundation of China (Nos. 11831015 and 61672388) , the National Key Research and Development Program of China (No. 2018YFC1314600) and the Natural Science Foundation of Hubei Province No. 2019CFA007.\vspace*{-12pt}

\bibliographystyle{natbib}
\bibliography{document}

\end{document}